\documentclass[runningheads]{llncs}
\usepackage{graphicx, upgreek, tikz, standalone}
\usetikzlibrary{arrows, decorations, graphs, decorations.pathmorphing, decorations.pathreplacing, decorations.shapes, backgrounds, positioning, fit, petri, bending, intersections, automata, shapes.misc, calc, arrows.meta}
\tikzset{}

\usepackage[utf8]{inputenc}

\begin{document}
\title{A Tsetlin Machine with Multigranular Clauses}
%
%
\author{Saeed Rahimi Gorji \inst{1}\orcidID{0000-0002-2699-9903} \and
Ole-Christoffer	Granmo\inst{1}\orcidID{0000-0002-7287-030X} \and
Adrian Phoulady\orcidID{0000-0002-4979-5751} \and
Morten Goodwin\inst{1}\orcidID{0000-0001-6331-702X}}
\authorrunning{S. R. Gorji et al.}
%
\institute{Centre for Artificial Intelligence Research, University of Agder, Grimstad, Norway
\email{\{saeed.r.gorji,ole.granmo,morten.goodwin\}@uia.no}}
%
\maketitle 
\begin{abstract}
The recently introduced Tsetlin Machine (TM) has provided competitive pattern recognition accuracy in several benchmarks, however, requires a 3-dimensional hyperparameter search. In this paper, we introduce the Multigranular Tsetlin Machine (MTM). The MTM eliminates the \emph{specificity} hyperparameter, used by the TM to control the granularity of the conjunctive clauses that it produces for recognizing patterns. Instead of using a fixed global specificity, we encode varying specificity as part of the clauses, rendering the clauses multigranular.   This makes it easier to configure the TM because the dimensionality of the hyperparameter search space is reduced to only two dimensions. Indeed, it turns out that there is significantly less hyper-parameter tuning involved in applying the MTM to new problems.  Further, we demonstrate empirically that the MTM provides similar performance to what is achieved with a finely specificity-optimized TM, by comparing their performance on both synthetic and real-world datasets.

\end{abstract}

\section{Introduction}
The Tsetlin Machine (TM) is a new machine learning algorithm that was introduced in 2018 \cite{granmo2018tsetlin}. It leverages the ability of so-called learning automata (LA) to learn the optimal action in unknown stochastic environments \cite{narendra1989learning}. The TM has provided competitive pattern recognition accuracy in several benchmarks, without losing the important property of interpretability \cite{granmo2018tsetlin}.

The TM builds upon a long tradition of LA research, involving cooperating systems of LA \cite{thathachar1987learning,sastry1999learning,sastry2009team,rahnamazadeh2017node}. More recently, LA have been combined with cellular automata (CA), where each CA cell contains one or more LA, which learn in a distributed fashion \cite{esmaeilpour2012cellular,ahangaran2017associative,uzun2018solution}. Some noteworthy LA-based classifiers are further introduced in \cite{barto1985pattern,goodwin2016distributed,afshar2013presenting,zahiri2012classification,aghaebrahimi2009data,zahiri2008learning}. However, these approaches mainly tackle small-scale pattern classification problems.

In all brevity, a TM consists of $m$ teams of Tsetlin Automata (TA) \cite{tsetlin1961behaviour} that interact to solve complex pattern recognition problems. It takes a binary feature vector $X=[x_1,x_2,...,x_n] \in \{0,1\}^n$ as input, which is further processed by $m$ conjunctive clauses $C^+_1, \ldots, C^+_{\frac{m}{2}}$ and $C^-_1 \ldots, C^-_{\frac{m}{2}}$. Each clause captures a specific sub-pattern, formulated as a conjunction of literals (binary features and their negations): $x_a \land \ldots \land x_b \land \lnot x_c \land \ldots \land \lnot x_d$. Half of the clauses are assigned positive polarity. These describe sub-patterns for output $y=1$. The other half is assigned negative polarity, describing sub-patterns for output $y=0$. The output $y \in \{0, 1\}$ is thus simply decided by a majority vote: $y = \left(\sum C^+_j - \sum C^-_j \ge 0\right)$.

During learning, each team of TA is responsible for a specific clause. There are two TA per feature $x_i$. One  decides whether to include $x_i$ in the clause, while the other decides upon including $\neg x_i$. These decisions are updated based on reinforcement derived from training examples $(\widehat{X}, \hat{y})$, contrasting the current clauses against $(\widehat{X}, \hat{y})$ (see \cite{granmo2018tsetlin} for further details).

Learning in the TM is governed by three hyperparameters: number of clauses $m$, specificity $s$, and voting target $T$, all set by the user \cite{granmo2018tsetlin}. The number of clauses $m$ decides the overall capacity of the TM to represent patterns, with each clause capturing a particular facet of the data. Specificity $s$, in turn, is used by the TM to control the granularity of the  clauses, playing a similar role as so-called \emph{support} in frequent itemset mining. Finally, the voting target $T$ produces an ensemble effect by stimulating up to $T$ clauses to output $1$ for each input, but not more than $T$. This drives the $m$ clauses to distribute themselves uniformly across the patterns present in the data, avoiding local optima.  In this paper, we will divide $T$ by the number of clauses $m$, to obtain a target value relative to the number of clauses.

\section{A Tsetlin Machine with multigranular clauses}

We now introduce the Multigranular Tsetlin Machine (MTM) with the goal of eliminating specificity $s$ as hyperparameter. Specificity controls how fine-grained patterns the TM seeks, and it is thus crucial to set this parameter correctly to maximize the accuracy of the resulting classifier. A poor choice for $s$ can easily result in inferior accuracy.

While $s$ is a global hyperparameter for the TM, to be set by the user, the MTM instead assigns a unique $s_j$-value local to each clause $C_j$, $1 \le j \le m$. In all brevity, we define a fixed range $[l, u]$ for $s_j$ and then assign $s_j$ a value decided by the clause index $j$:
\[
 s_j = (u-l) \cdot \frac{m-j}{m-1} + l.
\]
As seen, specificity values $\{s_j\}$ are decreasing linearly with the clause index $j$. In this paper, we use the range $l=2.0$ to $u=200.0$, which covers a wide range of both coarse and very fine patterns, as this range performs robustly across all of our experiments.

The above multigranular approach has two crucial effects. First, one avoids the need for finding a suitable value for $s$. Experimenting with different $s$-values can be computationally expensive, in particular for large datasets. Secondly, patterns of diverse frequencies can more easily be captured by the clauses when the clauses themselves reflect the diversity of the patterns. Indeed, the classic TM may in the worst case spend an unnecessary large amount of clauses to capture frequent patterns, when $s$ has been set to also capture less frequent patterns. This in turn may potentially clutter some clauses with unnecessary literals, making them less readable (of course, these unnecessary literals may also be pruned in a post-processing phase, but at a higher computational cost during learning). As an example, assume the classic TM tries to capture the pattern $x_1 \wedge \neg x_2$ of frequency $\frac{1}{4}$, with an $s$-value of $16$. In this case, the TM will potentially add two extra literals to the target pattern, introducing e.g. $x_1 \wedge \neg x_2 \wedge x_3 \wedge x_4$. Now, to capture the pattern $x_1 \wedge \neg x_2$, the TM must spend four clauses instead of one, that is, one clause per value configuration of  $x_3$ and $x_4$.

\section{Experimental results}
In this section, we present experimental results examining how multigranular clauses affect accuracy and learning speed, in comparison with the classic TM. For the classic TM algorithm, we used a grid search to find the best $s$-values as well as the threshold parameters. For MTM, however, we only needed to find an appropriate threshold value, using $s$-values in the form of an arithmetic progression from $2$ to $200$.

In our first experiment, we consider a problem that intermixes two kinds of patterns of different complexity. In brief, we specify patterns using $6$ binary variables $x_1, x_2, \ldots, x_6$. The patterns for output $y=0$ are either $\neg x_1 \wedge \neg x_2$ or the more elaborate $x_1 \wedge \neg (x_3 \oplus x_4 \oplus x_5 \oplus x_6 \oplus x_7)$, while the patterns for output $y=1$ are either $\neg x_1 \wedge x_2$ or $x_1 \wedge (x_3 \oplus x_4 \oplus x_5 \oplus x_6 \oplus x_7)$.

\begin{center}
\begin{table}
\caption{Accuracy after 200 and 500 epochs for TM and MTM on artificial data.}
\label{tab1}
\begin{tabular}{|c|c|c|c|c||c|c|c|}
\hline
Clauses & s & Threshold & TM (200) & TM (500) 
& Threshold & MTM (200) & MTM (500) \\
\hline
10 & 110 & 0.1 & 75.7\% & 78.2\% & 0.16 & 76.1\% & 78.0\% \\
20 & 100 & 0.06 & 76.6\% & 78.2\% & 0.08 & 78.8\% & 78.4\% \\
50 & 50 & 0.04 &  88.4\% & 89.2\% & 0.04 & 88.5\% & 88.2\% \\
100 & 60 & 0.03 &  94.3\% & 95.9\% & 0.02 & 93.2\% & 95.2\% \\
500 & 35 & 0.01 &  97.8\% & 98.0\% & 0.01 & 98.0\% & 98.0\% \\
\hline
\end{tabular}
\end{table}
\end{center}

Both the training and test sets consist of 300 randomly generated examples and approximately $25\%$ of the examples fall within each of the four patterns. Table \ref{tab1} shows the final accuracy for the TM and the MTM after 200 and 500 epochs, averaged over $10$ independent experiment runs, alongside the hyper-parameter values that led to that result. As seen, both algorithms exhibit similar performances for different number of clauses, however, MTM did not require tuning of $s$.

\begin{figure}
    \centering
    \begin{minipage}{0.48\textwidth}
        \centering
        \includegraphics[width=\textwidth]{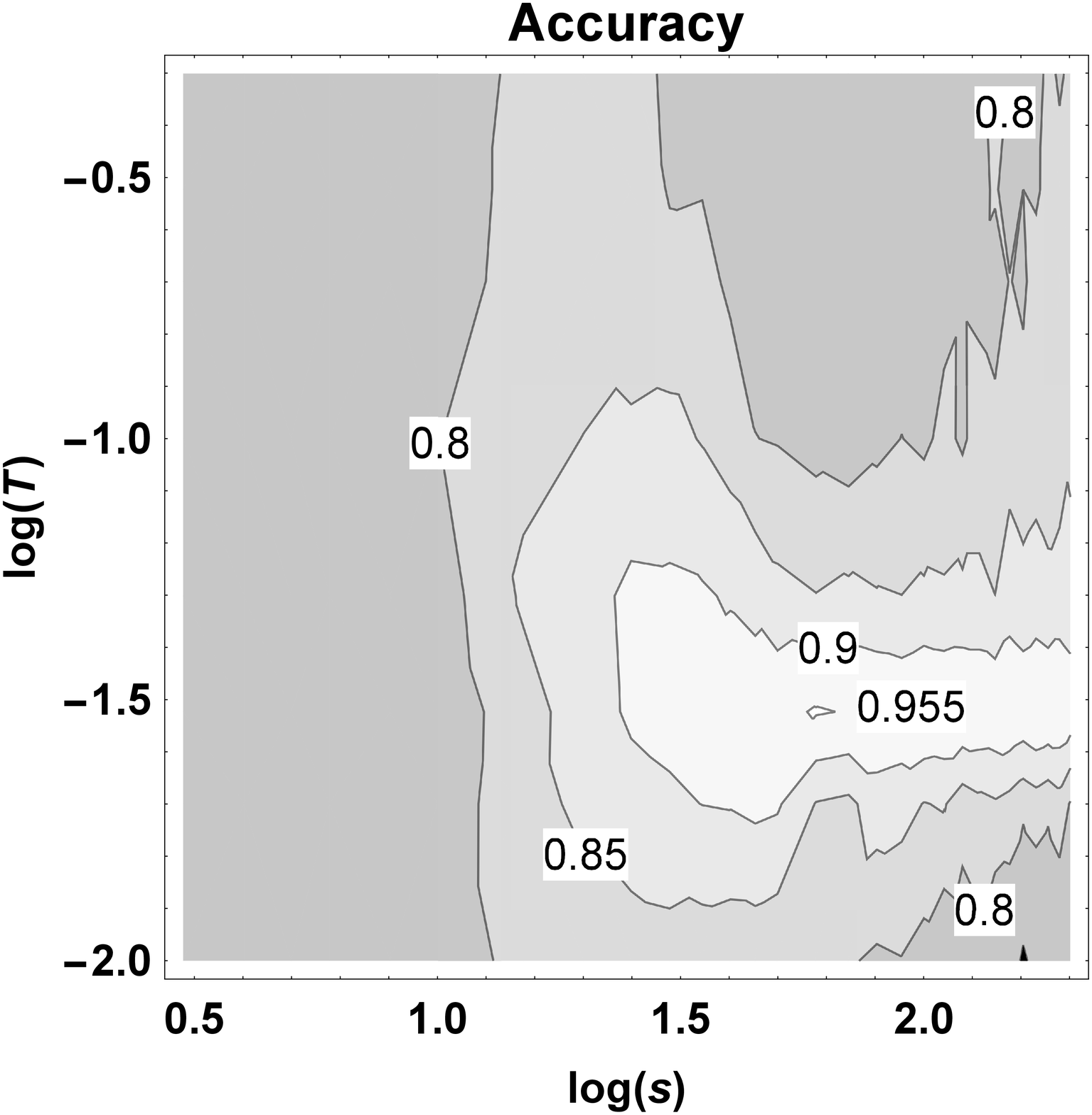}
        \caption{The Tsetlin machine's performance with 100 clauses after 500 epochs}
          \label{fig:cls100-xor15-0}
    \end{minipage}\hfill
    \begin{minipage}{0.48\textwidth}
        \centering
        \includegraphics[width=\textwidth]{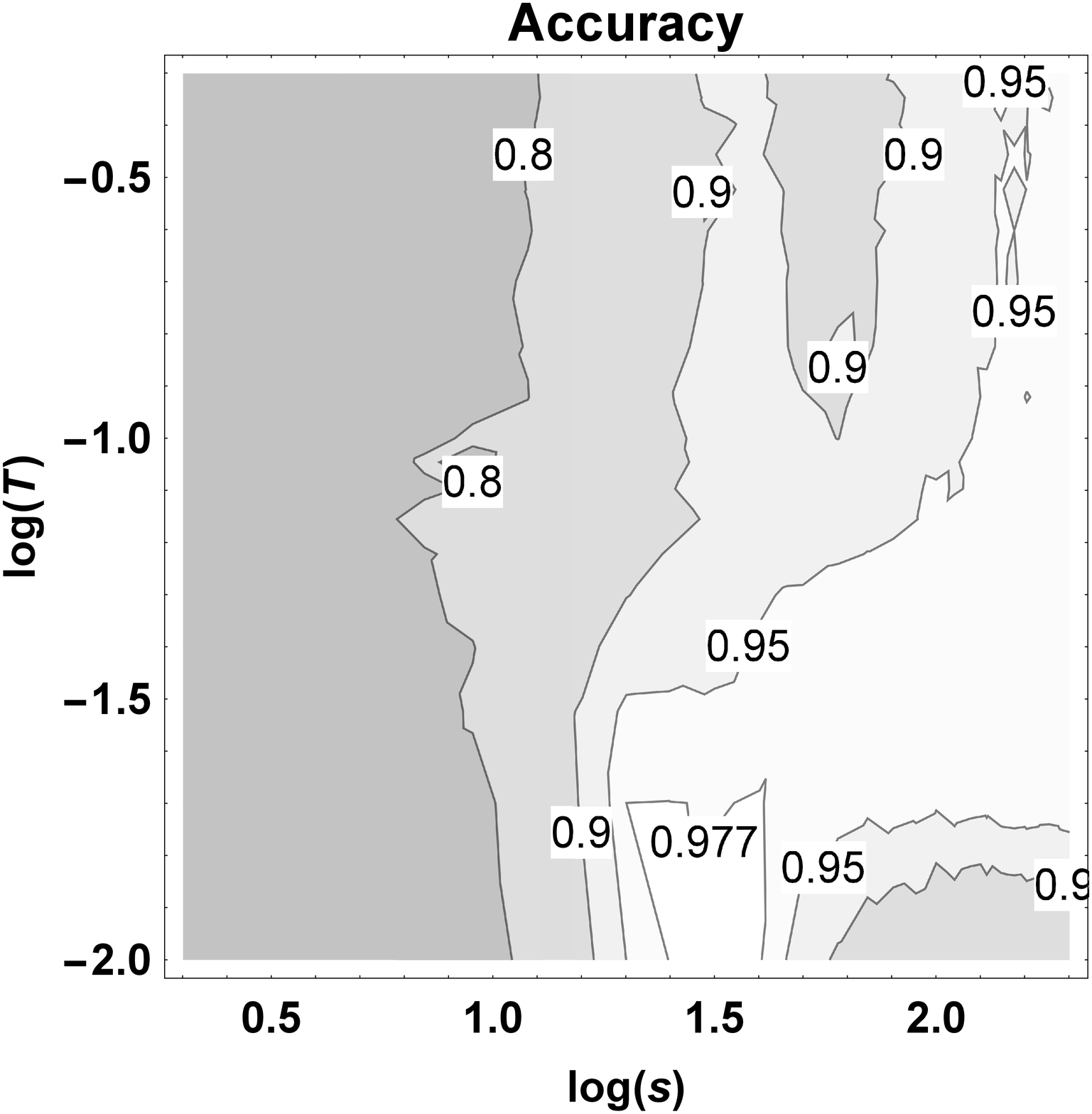}
        \caption{The Tsetlin machine's performance with 500 clauses after 500 epochs}
          \label{fig:cls500-xor15-0}
    \end{minipage}
\end{figure}

Fig. \ref{fig:cls100-xor15-0} and \ref{fig:cls500-xor15-0} depict accuracy as a function of the $s$ and the threshold parameters. As seen, finding high-performing hyperparameter values is not trivial, with the search space varying with the number of clauses employed. 
In contrast, MTM is optimized only with respect to the threshold.

\begin{figure}
    \centering
    \begin{minipage}{0.48\textwidth}
        \centering
        \includegraphics[width=\textwidth]{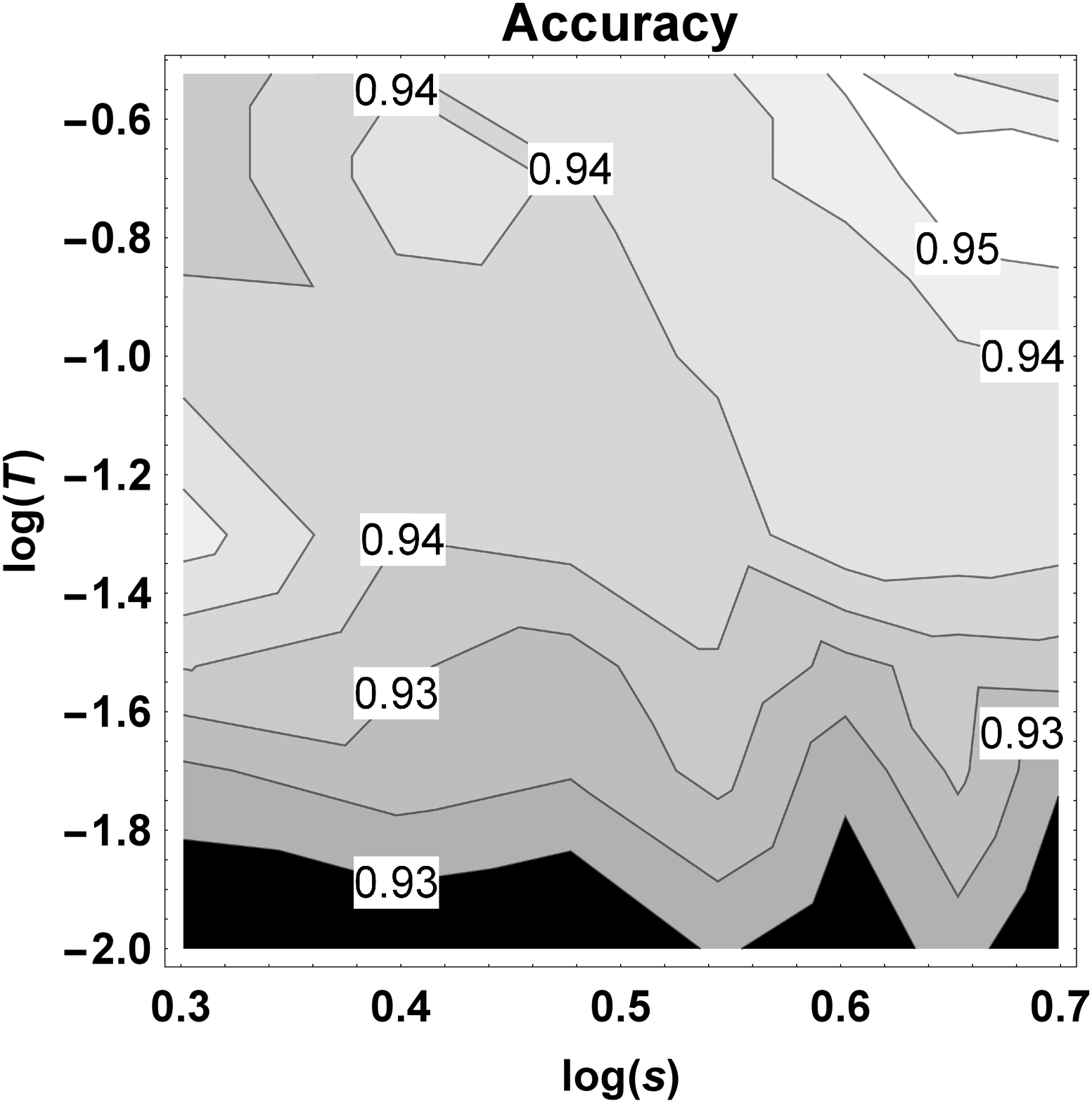}
        \caption{The Tsetlin machine's performance with 100 clauses after 500 epochs}
          \label{fig:cls100-iris20}
    \end{minipage}\hfill
    \begin{minipage}{0.48\textwidth}
        \centering
        \includegraphics[width=\textwidth]{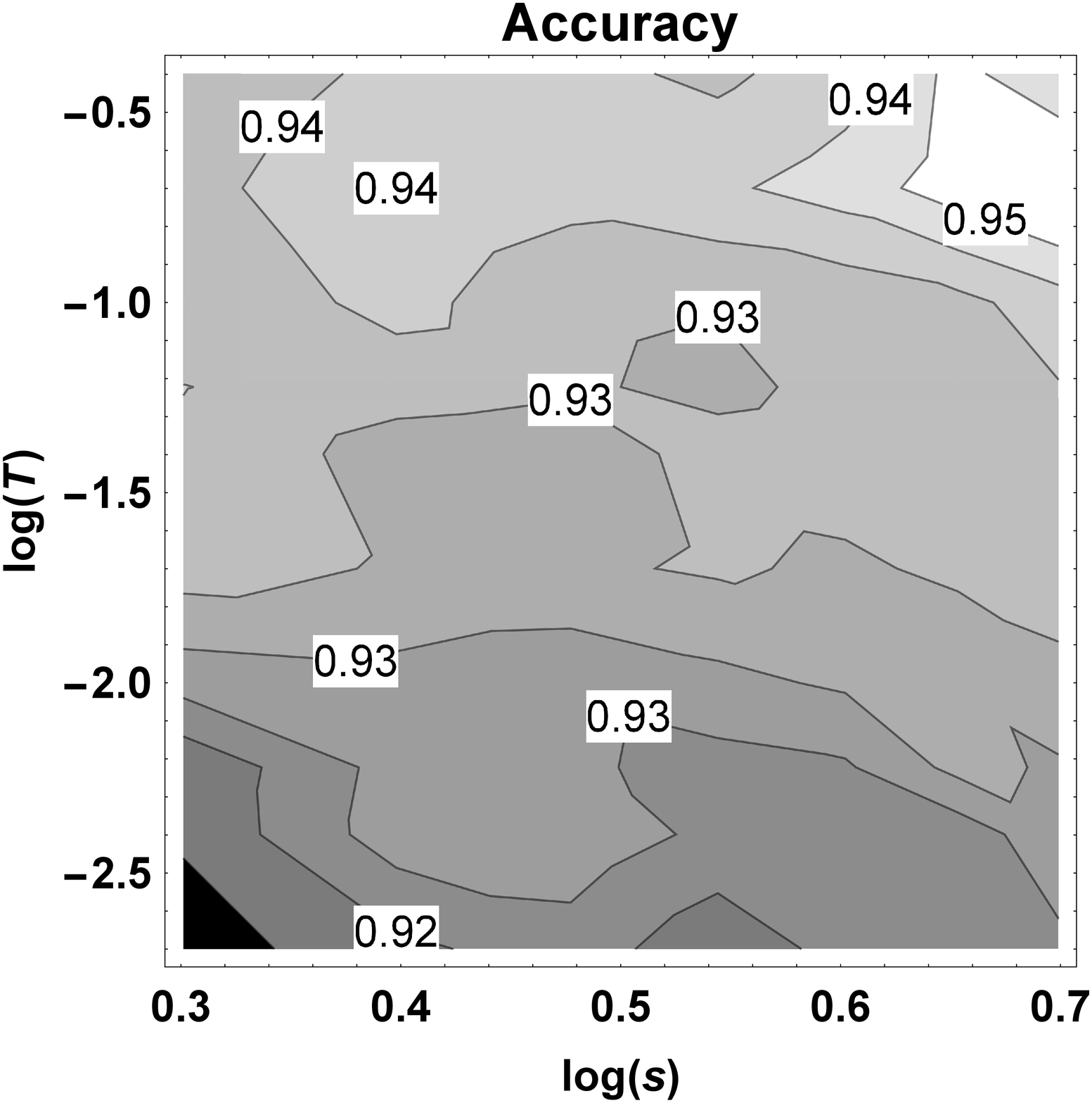}
        \caption{The Tsetlin machine's performance with 500 clauses after 500 epochs}
          \label{fig:cls500-iris20}
    \end{minipage}
\end{figure}

In our second experiment, we evaluate performance on the Iris flower dataset\footnote{https://archive.ics.uci.edu/ml/datasets/iris} \cite{Dua:2019}. 
This dataset contains measurements for three classes of flowers, 50 instances of each.  Each instance consists of four real-valued features. We used five bits to represent each real number (three and two bits for the integer and fractional parts, respectively). We further employed $10$ random 80\%-20\% training-test splits to increase the robustness of the evaluation. The results reported in Table \ref{tab2} are the average performance of $10$ independent experiment runs, for each training-test split. Fig. \ref{fig:cls100-iris20} and \ref{fig:cls500-iris20} capture the difficulty of finding suitable values for the hyperparameters, while Table \ref{tab2} shows how the MTM attains slightly lower accuracy compared to the classic TM, however, by only fine-tuning the threshold value.

\begin{table}
\caption{The accuracy of TM and MTM on the binary Iris dataset}
\label{tab2}
\begin{tabular}{|c|c|c|c|c|}
\hline
Epoch & TM (100 clauses) & TM (500 clauses) & MTM (100 clauses) & MTM (500 clauses) \\
 & $(s = 5, T = 0.2)$ & $(s = 5, T = 0.2)$ & $(T = 0.05)$ & $(T = 0.03)$ \\
\hline
100 & 95.1\% & 95.5\% & 94.2\% & 95.0\% \\
200 & 95.3\% & 95.6\% & 94.5\% & 94.6\% \\
300 & 95.1\% & 95.7\% & 94.5\% & 94.9\% \\
500 & 95.2\% & 95.7\% & 94.7\% & 95.0\% \\
\hline
\end{tabular}
\end{table}

\noindent Further experiments can be found in the unabridged version of this paper \cite{gorji2019}.

\section{Conclusion}
In this work we introduced the multigranular Tsetlin Machine (MTM) to reduce the complexity of the hyperparameter search in Tsetlin Machine (TM) based learning. We achieved this by eliminating the specificity hyperparameter $s$, instead introducing clauses with unique and diverse local $s$-values.
In our empirical results, it turns out that we actually can obtain similar accuracy as a finely optimized classic TM, however, eliminating the need to consider $s$.
Furthermore, we explored the capability of the MTM to capture patterns of diverse frequencies by using an artificial dataset.

As further research, a natural next step is to work on the theoretical aspects of MTM. Although the theoretical convergence results for TM also should hold for MTM, this needs to be investigated more rigorously. Furthermore, other interesting areas of research could be mechanisms for improving convergence speed. Finally, we intend to investigate the possibility of eliminating the other two remaining hyperparameters as well, making the TM completely parameter-free.

\bibliographystyle{splncs04}
\bibliography{ref}

\begin{thebibliography}{10}
\providecommand{\url}[1]{\texttt{#1}}
\providecommand{\urlprefix}{URL }
\providecommand{\doi}[1]{https://doi.org/#1}

\bibitem{afshar2013presenting}
Afshar, S., Mosleh, M., Kheyrandish, M.: Presenting a new multiclass classifier
  based on learning automata. Neurocomputing  \textbf{104},  97--104 (2013)

\bibitem{aghaebrahimi2009data}
Aghaebrahimi, M., Zahiri, S., Amiri, M.: Data mining using learning automata.
  World Acad. Sci. Eng. Technol  \textbf{49},  343--351 (2009)

\bibitem{ahangaran2017associative}
Ahangaran, M., Taghizadeh, N., Beigy, H.: Associative cellular learning
  automata and its applications. Applied Soft Computing  \textbf{53},  1--18
  (2017)

\bibitem{barto1985pattern}
Barto, A.G., Anandan, P.: Pattern-recognizing stochastic learning automata.
  IEEE Transactions on Systems, Man, and Cybernetics (3),  360--375 (1985)

\bibitem{Dua:2019}
Dua, D., Graff, C.: {UCI} machine learning repository (2017),
  \url{http://archive.ics.uci.edu/ml}

\bibitem{esmaeilpour2012cellular}
Esmaeilpour, M., Naderifar, V., Shukur, Z.: Cellular learning automata approach
  for data classification. International Journal of Innovative Computing,
  Information and Control  \textbf{8}(12),  8063--8076 (2012)

\bibitem{goodwin2016distributed}
Goodwin, M., Yazidi, A., Jonassen, T.M.: Distributed learning automata for
  solving a classification task. In: 2016 IEEE congress on evolutionary
  computation (CEC). pp. 3999--4006. IEEE (2016)

\bibitem{gorji2019}
Gorji, S.R., Granmo, O.C., Phoulady, A., Goodwin, M.: {A Tsetlin Machine with
  Multigranular Clauses and its Applications}. Unabridged journal version of
  this paper. To be submitted.  (2019)

\bibitem{granmo2018tsetlin}
Granmo, O.C.: {The Tsetlin Machine - A Game Theoretic Bandit Driven Approach to
  Optimal Pattern Recognition with Propositional Logic}. arXiv preprint
  arXiv:1804.01508  (2018)

\bibitem{narendra1989learning}
Narendra, K., Thathachar, M.: Learning Automata: An Introduction. Prentice-Hall
  International (1989), \url{https://books.google.no/books?id=ljphQgAACAAJ}

\bibitem{rahnamazadeh2017node}
Rahnamazadeh, A., Meybodi, M.R., Kadkhoda, M.T.: Node classification in social
  network by distributed learning automata. Information Systems \&
  Telecommunication p.~111 (2017)

\bibitem{sastry2009team}
Sastry, P., Nagendra, G., Manwani, N.: A team of continuous-action learning
  automata for noise-tolerant learning of half-spaces. IEEE Transactions on
  Systems, Man, and Cybernetics, Part B (Cybernetics)  \textbf{40}(1),  19--28
  (2009)

\bibitem{sastry1999learning}
Sastry, P., Thathachar, M.: Learning automata algorithms for pattern
  classification. Sadhana  \textbf{24}(4-5),  261--292 (1999)

\bibitem{thathachar1987learning}
Thathachar, M.A., Sastry, P.S.: Learning optimal discriminant functions through
  a cooperative game of automata. IEEE Transactions on Systems, Man, and
  Cybernetics  \textbf{17}(1),  73--85 (1987)

\bibitem{tsetlin1961behaviour}
Tsetlin, M.L.: On behaviour of finite automata in random medium. Avtom I
  Telemekhanika  \textbf{22}(10),  1345--1354 (1961)

\bibitem{uzun2018solution}
Uzun, A.O., Usta, T., D{\"u}ndar, E.B., Korkmaz, E.E.: A solution to the
  classification problem with cellular automata. Pattern Recognition Letters
  \textbf{116},  114--120 (2018)

\bibitem{zahiri2008learning}
Zahiri, S.H.: Learning automata based classifier. Pattern Recognition Letters
  \textbf{29}(1),  40--48 (2008)

\bibitem{zahiri2012classification}
Zahiri, S.H.: Classification rule discovery using learning automata.
  International Journal of Machine Learning and Cybernetics  \textbf{3}(3),
  205--213 (2012)

\end{thebibliography}

\end{document}